\documentclass[11pt,a4paper]{article}
\usepackage[hyperref]{emnlp-ijcnlp-2019}
\usepackage[utf8]{inputenc}
\usepackage{times}
\usepackage{latexsym}
\usepackage{url}
\usepackage{graphicx}
\usepackage{amsmath}
\usepackage{amsfonts}
\usepackage{booktabs}
\usepackage{multicol}
\usepackage{multirow}
\usepackage{tabularx}
\usepackage{color}
\usepackage[linesnumbered,ruled]{algorithm2e}
\DeclareMathOperator*{\argmax}{argmax}
\newcommand{\RNum}[1]{\uppercase\expandafter{\romannumeral #1\relax}}

\usepackage{xr}
\makeatletter
\newcommand*{\addFileDependency}[1]{
  \typeout{(#1)}
  \@addtofilelist{#1}
  \IfFileExists{#1}{}{\typeout{No file #1.}}
}
\makeatother

\aclfinalcopy 

\newcommand{\citesub}[1]{\citeauthor{#1} \shortcite{#1}}

\title{Guided Dialog Policy Learning: \\ Reward Estimation for Multi-Domain Task-Oriented Dialog}
\author{Ryuichi Takanobu$^1$, Hanlin Zhu$^2$, Minlie Huang$^1$\footnotemark[1]\\
Institute for AI, BNRist, $^1$DCST, $^2$IIIS, Tsinghua University, Beijing, China\\
{\tt gxly19@mails.tsinghua.edu.cn, aihuang@tsinghua.edu.cn}}

\begin{document}
\maketitle
\renewcommand{\thefootnote}{\fnsymbol{footnote}}
\footnotetext[1]{Corresponding author}
\renewcommand{\thefootnote}{\arabic{footnote}}
\begin{abstract}
    Dialog policy decides what and how a task-oriented dialog system will respond, and plays a vital role in delivering effective conversations.
    Many studies apply Reinforcement Learning to learn a dialog policy with the reward function which requires elaborate design and pre-specified user goals.
    With the growing needs to handle complex goals across multiple domains, such manually designed reward functions are not affordable to deal with the complexity of real-world tasks. To this end, we propose Guided Dialog Policy Learning, a novel algorithm based on Adversarial Inverse Reinforcement Learning for joint reward estimation and policy optimization in multi-domain task-oriented dialog. The proposed approach estimates the reward signal and infers the user goal in the dialog sessions. The reward estimator evaluates the state-action pairs so that it can guide the dialog policy at each dialog turn. Extensive experiments on a multi-domain dialog dataset show that the dialog policy guided by the learned reward function achieves remarkably higher task success than state-of-the-art baselines.
\end{abstract}

\section{Introduction}
Dialog policy, deciding the next action that the dialog agent should take at each turn, is a crucial component of a task-oriented dialog system. Among many models,  Reinforcement Learning (RL) is commonly used to learn dialog policy \cite{fatemi2016policy,peng2017composite,chen2017agent,yarats2018hierarchical,lei2018sequicity,he2018decoupling,su2018discriminative}, where users are modeled as a part of the environment and the policy is learned through interactions with users.

While it is too expensive to learn directly from real users since RL requires a large number of samples to train, most existing studies use data-driven approaches to build a dialog system from conversational corpora \cite{zhao2016towards,dhingra2017towards,shah2018bootstrapping,shi2018sentiment}, where a common strategy is to build a user simulator, and then to learn dialog policy through making simulated interactions between an agent and the simulator. A typical reward function on policy learning consists of a small negative penalty at each turn to encourage a shorter session, and a large positive reward when the session ends successfully if the agent completes the user goal.

\begin{table}[!tb]
    \centering
    \small
    \begin{tabular}[\linewidth]{l@{~}l}
        \toprule
        U: & I'm looking for a hotel to stay that has 5 stars and \\ & cheap price range. \\
        S: & I am sorry that there is no such hotel, would you like \\ & to reserve a 3-star hotel as an alternative? \\
        U: & I'd prefer a 4-star hotel even if it's a bit expensive. \\ & Oh, and I need parking. \\
        S: & OK, I find a moderately priced 4-star hotel that \\ & includes parking and free wifi. \\
        U: & Are there any places to eat around it? \\
        S: & Many. Japanese, Indian, French, etc. What kind of \\ & food would you like? \\
        \bottomrule
    \end{tabular}
    \caption{An example of the multi-domain task-oriented dialog between the user (U) and the system (S).
    The dialog proceeds successfully because the system informs the user that no matching hotel exists (the first turn), identifies the new user goal about parking (the second turn), and shifts the topic to the restaurant domain (the third turn), which well understands the user's demand.
    }
    \label{tab:example}
\end{table}

However, specifying an effective reward function is challenging in task-oriented dialog. On one hand, the short dialogs resulted from the negative constant rewards are not always efficient. The agent may end a session too quickly to complete the task properly. For example, it is inappropriate to book a 3-star hotel without confirming with the user at the first turn in Table \ref{tab:example}.
On the other hand, an explicit user goal is essential to evaluate the task success in the reward design, but user goals are hardly available in real situations \cite{su2016line}. In addition, the user goal may change as the conversation proceeds. For instance, the user introduces a new requirement for the parking information at the second turn in Table \ref{tab:example}.

Unlike a handcrafted reward function that only evaluates the task success at the end of a session, a good reward function should be able to guide the policy dynamically to complete the task during the conversation. We refer to this as the \textit{reward sparsity} issue.
Furthermore, the reward function is often manually tweaked until the dialog policy performs desired behaviors. With the growing needs for the system to handle complex tasks across multiple domains, a more sophisticated reward function would be designed, which poses a serious challenge to manually trade off those different factors.

In this paper, we propose a novel model for learning task-oriented dialog policy. The model includes a robust dialog reward estimator based on Inverse Reinforcement Learning (IRL).
The main idea is to automatically infer the reward and goal that motivates human behaviors and interactions from the real human-human dialog sessions.
Different from conventional IRL that learns a reward function first and then trains the policy, we integrate Adversarial Learning (AL) into the method so that the policy and reward estimator can be learned simultaneously in an alternate way, thus improving each other during training. To deal with reward sparsity, the reward estimator evaluates the generated dialog session using state-action pairs instead of the entire session, which provides reward signals at each dialog turn and guides dialog policy learning better.

To evaluate the proposed approach, we conduct our experiments on a multi-domain, multi-intent task-oriented dialog corpus. The corpus involves large state and action spaces, multiple decision making in one turn, which makes it more challenging for the reward estimator to infer the user goal. Furthermore, we experiment with two different user simulators.
The contributions of our work are in three folds:
\begin{itemize}
    \item We build a reward estimator via Inverse Reinforcement Learning (IRL) to infer an appropriate reward from multi-domain dialog sessions, in order to avoid manual design of reward function.

    \item We integrate Adversarial Learning (AL) to train the policy and estimator simultaneously, and evaluate the policy using state-action pairs to better guide dialog policy learning.

    \item We conduct experiments on the multi-domain, multi-intent task-oriented dialog corpus, with different types of user simulators. Results show the superiority of our model to the state-of-the-art baselines.
\end{itemize}

\section{Related Work}
\subsection{Multi-Domain Dialog Policy Learning}
Some recent efforts have been paid to multi-domain task-oriented dialog systems where users converse with the agent across multiple domains.
A natural way to handle multi-domain dialog systems is to learn multiple independent single-domain sub-policies \cite{wang2014policy,gavsic2015policy,cuayahuitl2016deep}. Multi-domain dialog completion was also addressed by hierarchical RL which decomposes the task into several sub-tasks in terms of temporal order \cite{peng2017composite} or space abstraction \cite{casanueva2018feudal}, but the hierarchical structure can be very complex and constraints between different domains should be considered if an agent conveys multiple intents.

\subsection{Reward Learning in Dialog Systems}
Handcrafted reward functions for dialog policy learning require elaborate design. Several reward learning algorithms have been proposed to find better rewards, including supervised learning on expert dialogs \cite{li2014temporal}, online active learning from user feedback \cite{su2016line}, multi-object RL to aggregate measurements of various aspects of user satisfaction \cite{ultes2017reward}, etc. However, these methods still require some knowledge about user goals or annotations of dialog ratings from real users. \citesub{boularias2010learning} and \citesub{rojas2014bayesian} learn the reward from dialogs using linear programming based on IRL, but do not scale well in real applications. Recently, \citesub{liu2018adversarial} use adversarial rewards as the only source of reward signal. It trains a Bi-LSTM as a discriminator that works on the entire session to predict the task success.

\subsection{Adversarial Inverse Reinforcement Learning}
IRL aims to infer the reward function $\mathcal{R}$ underlying expert demonstrations sampled from humans or the optimal policy $\pi^*$. This is similar to the discriminator network in AL that evaluates how realistic the sample looks. \citesub{finn2016connection} draw a strong connection between GAN and maximum entropy causal IRL \cite{ziebart2010modeling} by replacing the estimated data density in AL with the Boltzmann distribution in IRL, i.e. $p(x) \propto \exp(-E(x))$. Several approaches \cite{ho2016generative,fu2018learning} obtain a promising result on automatic reward estimation in large, high-dimensional environments by combining AL with IRL. Inspired by this, we apply AIRL to complex, multi-domain task-oriented dialog, which faces new issues such as discrete action space and language understanding.

\section{Guided Dialog Policy Learning}\label{sec:GDPL}
We propose Guided Dialog Policy Learning (GDPL), a flexible and practical method on joint reward learning and policy optimization for multi-domain task-oriented dialog systems.

\begin{figure}[!tb]
    \centering
    \includegraphics[width=\linewidth]{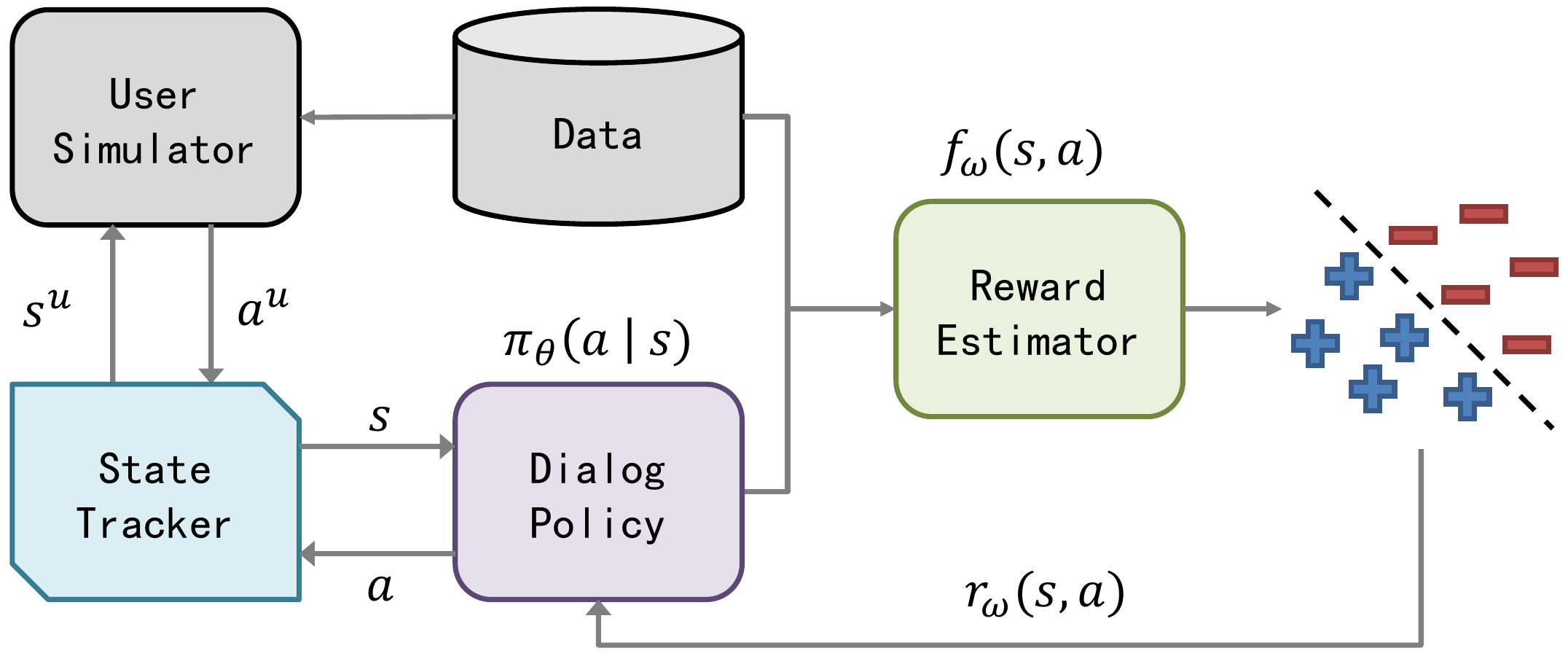}
    \caption{Architecture of GDPL. The dialog policy $\pi$ decides the dialog act $a$ according to the dialog state $s$ provided by the state tracker, and the reward estimator $f$ evaluates the dialog policy by comparing the generated state-action pair $(s, a)$ with the human dialog.
    }
    \label{fig:framework}
\end{figure}

\subsection{Overview}
The overview of the full model is depicted in Fig. \ref{fig:framework}. The framework consists of three modules: a multi-domain Dialog State Tracker (DST) at the dialog act level, a dialog policy module for deciding the next dialog act, and a reward estimator for policy evaluation.

Specifically, given a set of collected human dialog sessions $\mathcal{D} = \{\tau_1, \tau_2, \dots \}$, each dialog session $\tau$ is a trajectory of state-action pairs $\{s_0^u, a_0^u, s_0, a_0, s_1^u, a_1^u, s_1, a_1, \dots\}$. The user simulator $\mu(a^u, t^u|s^u)$ posts a response $a^u$ according to the user dialog state $s^u$ where $t^u$ denotes a binary terminal signal indicating whether the user wants to end the dialog session. The dialog policy $\pi_\theta(a|s)$ decides the action $a$ according to the current state $s$ and interacts with the simulator $\mu$. During the conversation, DST records the action from one dialog party and returns the state to the other party for deciding what action to take in the next step. Then, the reward estimator $f_\omega(s, a)$ evaluates the quality of the response from the dialog policy, by comparing it with sampled human dialog sessions from the corpus. The dialog policy $\pi$ and the reward estimator $f$ are MLPs parameterized by $\theta, \omega$ respectively.
Note that our approach does not need any human supervision during training, and modeling a user simulator is beyond the scope of this paper.

In the subsequent subsections, we will first explain the state, action, and DST used in our algorithm. Then, the algorithm is introduced in a session level, and last followed by a decomposition of state-action pair level.

\subsection{Multi-Domain Dialog State Tracker}
A dialog state tracker keeps track of the dialog session to update the dialog state \cite{williams2016dialog,zhang2019memory}. It records \textit{informable} slots about the constraints from users and \textit{requestable} slots that indicates what users want to inquiry. DST maintains a separate \textit{belief state} for each slot. Given a user action, the belief state of its slot type is updated according to its slot value \cite{roy2000spoken}. Action and state in our algorithm are defined as follows:

\paragraph{Action}: Each system action $a$ or user action $a^u$ is a subset of dialog act set $\mathcal{A}$ as there may be multiple intents in one dialog turn. A \textit{dialog act} is an abstract representation of an intention \cite{stolcke2000dialogue}, which can be represented in a quadruple composed of \textit{domain}, \textit{intent}, \textit{slot type} and \textit{slot value} in the multi-domain setting (e.g. [\textit{restaurant}, \textit{inform}, \textit{food}, \textit{Italian}]). In practice, dialog acts are delexicalized in the dialog policy. We replace the slot value with a count placeholder and refill it with the true value according to the entity selected from the external database, which allows the system to operate on unseen values.

\paragraph{State}: At dialog turn $t$\footnote{We regard a user turn and a system turn as one dialog turn throughout the paper.}, the system state $s_t = [a_t^u;a_{t-1};b_t;q_t]$ consists of (\RNum{1}) user action at current turn $a_t^u$; (\RNum{2}) system action at the last turn $a_{t-1}$; (\RNum{3}) all belief state $b_t$ from DST; and (\RNum{4}) embedding vectors of the number of query results $q_t$ from the external database.

As our model works at the dialog act level, DST can be simply implemented by extracting the slots from actions.

\subsection{Session Level Reward Estimation}
Based on maximum entropy IRL \cite{ziebart2008maximum}, the reward estimator maximizes the log likelihood of observed human dialog sessions to infer the underlying goal,
\begin{equation*}
\begin{aligned}
    \omega^* &= \argmax_{\omega} \mathbb{E}_{\tau \sim \mathcal{D}} [f_\omega(\tau)], \\
    f_\omega(\tau) &= \log p_\omega(\tau) = \log \dfrac{e^{R_\omega(\tau)}}{Z_\omega}, \\
    R_\omega(\tau) &= \sum_{t=0}^{T} \gamma^t r_\omega(s_t, a_t), \\
    Z_\omega &= \sum_\tau e^{R_\omega(\tau)}.
\end{aligned}
\end{equation*}
where $f$ models human dialogs as a Boltzmann distribution \cite{ziebart2008maximum}, $R$ stands for the \textit{return} of a session, i.e. $\gamma$-discounted cumulative rewards, and $Z$ is the corresponding partition function.

The dialog policy is encouraged to mimic human dialog behaviors. It maximizes the expected entropy-regularized return $\mathbb{E}_\pi[R] + H(\pi)$ \cite{ziebart2010modeling} based on the \textit{principle of maximum entropy} through minimizing the KL-divergence between the policy distribution and Boltzmann distribution,
\begin{equation*}
\begin{aligned}
    J_\pi(\theta) &= - KL(\pi_\theta(\tau) || p_\omega(\tau))  \\
    &= \mathbb{E}_{\tau \sim \pi}[f_\omega(\tau) - \log \pi_\theta(\tau)] \\
    &= \mathbb{E}_{\tau \sim \pi}[R_\omega(\tau)] - \log Z_\omega + H(\pi_\theta),
\end{aligned}
\end{equation*}
where the term $\log Z_\omega$ is independent to $\theta$, and $H(\cdot)$ denotes the entropy of a model. Intuitively, maximizing entropy is to resolve the ambiguity of language that many optimal policies can explain a set of natural dialog sessions. With the aid of the likelihood ratio trick, the gradient for the dialog policy is
\begin{equation*}
\begin{aligned}
    \nabla_{\theta} J_\pi = \mathbb{E}_{\tau \sim \pi}[(f_\omega(\tau) - \log \pi_\theta(\tau)) \nabla_{\theta} \log \pi_\theta(\tau)].
\end{aligned}
\end{equation*}

In the fashion of AL, the reward estimator aims to distinguish real human sessions and generated sessions from the dialog policy. Therefore, it minimizes KL-divergence with the empirical distribution, while maximizing the KL-divergence with the policy distribution,
\begin{equation*}
\begin{aligned}
    J_f(\omega) &\hspace{-0.2em}=\hspace{-0.2em}-KL(p_\mathcal{D}(\tau) || p_\omega(\tau))\hspace{-0.2em}+\hspace{-0.2em}KL(\pi_\theta(\tau) || p_\omega(\tau)) \\
    &\hspace{-1em}=\hspace{-0.2em}\mathbb{E}_{\tau \sim \mathcal{D}}[f_\omega(\tau)]\hspace{-0.2em}+\hspace{-0.2em}H(p)\hspace{-0.2em}-\hspace{-0.2em}\mathbb{E}_{\tau \sim \pi}[f_\omega(\tau)]\hspace{-0.2em}-\hspace{-0.2em}H(\pi_\theta).
\end{aligned}
\end{equation*}
Similarly, $H(p)$ and $H(\pi_\theta)$ is independent to $\omega$, so the gradient for the reward estimator yields
\begin{equation*}
\begin{aligned}
    \nabla_{\omega} J_f = \mathbb{E}_{\tau \sim \mathcal{D}}[\nabla_{\omega} f_\omega(\tau)] - \mathbb{E}_{\tau \sim \pi}[\nabla_{\omega} f_\omega(\tau)].
\end{aligned}
\end{equation*}

\subsection{State-Action Level Reward Estimation}
So far, the reward estimation uses the entire session $\tau$, which can be very inefficient because of reward sparsity and may be of high variance due to the different lengths of sessions. Here we decompose a session $\tau$ into state-action pairs $(s, a)$ in the reward estimator to address the issues. Therefore, the loss functions for the dialog policy and the reward estimator become respectively as follows:
\begin{align}
    J_\pi(\theta) &=\mathbb{E}_{s,a \sim \pi}[\sum_{k=t}^T \gamma^{k-t} (f_\omega(s_k, a_k) \notag \\
    & \quad \quad \quad \quad - \log \pi_\theta(a_k|s_k))], \\
    J_f(\omega) &=\mathbb{E}_{s,a \sim \mathcal{D}}[f_\omega(s, a)] - \mathbb{E}_{s,a \sim \pi}[f_\omega(s, a)], \label{eq:estimator}
\end{align}
where $T$ is the number of dialog turns. Since the reward estimator evaluates a state-action pair, it can guide the dialog policy at each dialog turn with the predicted reward $\hat{r}_\omega(s,a) = f_\omega(s, a) - \log \pi_\theta(a|s)$.

\begin{algorithm}[!tb]
\DontPrintSemicolon
\SetKwInOut{Input}{Require}
\Input{Dialog corpus $\mathcal{D}$, User simulator $\mu$}
\ForEach{training iteration}{
    Sample human dialog sessions $D_H$ from $\mathcal{D}$ randomly \\
    Collect the dialog sessions $D_\Pi$ by executing the dialog policy $\pi$ and interacting with $\mu$, $a^u \sim \mu(\cdot|s^u)$, $a \sim \pi(\cdot|s)$, where $s$ is maintained by DST \\
    Update the reward estimator $f$ by maximizing $J_f$ w.r.t. $\omega$ adversarially (Eq. \ref{eq:estimator}) \\
    Compute the estimated reward of each state-action pair in $D_\Pi$, $\hat{r} = f_\omega(s,a) - \log \pi_\theta(a|s)$ \\
    Update $\pi$, $V$ using the estimated reward $\hat{r}$ by maximizing $J_\pi$, $J_V$ w.r.t. $\theta$ (Eq. \ref{eq:policy} and Eq. \ref{eq:value})\\
}
\caption{Guided Dialog Policy Learning}
\label{algorithm}
\end{algorithm}

Moreover, the reward estimator $f_\omega$ can be transformed to a reward approximator $g_\omega$ and a shaping term $h_\omega$ according to \cite{fu2018learning} to recover an interpretable and robust reward from real human sessions. Formally,
\begin{equation*}
    f_\omega(s_t, a_t, s_{t+1}) = g_\omega(s_t, a_t) + \gamma h_\omega(s_{t+1}) - h_\omega(s_t),
\end{equation*}
where we replace the state-action pair $(s_t, a_t)$ with the state-action-state triple $(s_t, a_t, s_{t+1})$ as the input of the reward estimator. Note that, different from the objective in \cite{fu2018learning} that learns a discriminator in the form $D_\omega(s,a) = \frac{p_\omega(s,a)}{p_\omega(s,a) + \pi(a|s)}$,
GDPL directly optimizes $f_\omega$, which avoids unstable or vanishing gradient issue in vanilla GAN \cite{arjovsky2017wasserstein}.

In practice, we apply Proximal Policy Optimization (PPO) \cite{schulman2017proximal}, a simple and stable policy based RL algorithm using a constant clipping mechanism as the soft constraint for dialog policy optimization,
\begin{align}
    J_\pi(\theta) &\hspace{-0.2em}=\hspace{-0.2em}\mathbb{E}_{s,a \sim \pi}[\min\{\beta_t \hat{A}_t, clip(\beta_t,\hspace{-0.2em}1 \hspace{-0.2em}-\hspace{-0.2em} \epsilon,\hspace{-0.2em}1 \hspace{-0.2em}+\hspace{-0.2em} \epsilon) \hat{A}_t\}], \label{eq:policy} \\
    \hat{A}_t &\hspace{-0.2em}=\hspace{-0.2em} \delta_t + \gamma \lambda \hat{A}_{t+1},  \notag \\
    \delta_t &\hspace{-0.2em}=\hspace{-0.2em} \hat{r}_t + \gamma V_\theta(s_{t+1}) - V_\theta(s_t), \notag \\
    J_V(\theta) &\hspace{-0.2em}=\hspace{-0.2em} - (V_\theta(s_t) - \sum_{k=t}^T \gamma^{k-t} \hat{r}_k)^2, \label{eq:value}
\end{align}
where $V_\theta$ is the approximate value function, $\beta_t = \frac{\pi_\theta(a_t|s_t)}{\pi_{\theta_{old}}(a_t|s_t)}$ is the ratio of the probability under the new and old policies, $\hat{A}$ is the estimated advantage, $\delta$ is TD residual, $\lambda$ and $\epsilon$ are hyper-parameters.

In summary, a brief script for GPDL algorithm is shown in Algorithm \ref{algorithm}.

\section{Experimental Setting}

\subsection{Data and Simulators}
We use MultiWOZ \cite{budzianowski2018multiwoz}, a multi-domain, multi-intent task-oriented dialog corpus that contains 7 domains, 13 intents, 25 slot types, 10,483 dialog sessions, and 71,544 dialog turns in our experiments. Among all the sessions, 1,000 each are used for validation and test. During the data collection process, a user is asked to follow a pre-specified user goal, but it encourages the user to change its goal during the session and the changed goal is also stored, so the collected dialogs are much closer to reality. The corpus also provides the ontology that defines all the entity attributes for the external database.

We apply two user simulators as the interaction environment for the agent. One is the agenda-based user simulator \cite{schatzmann2007agenda} which uses heuristics, and the other is a data-driven neural model, namely, Variational Hierarchical User Simulator (VHUS) derived from \cite{gur2018user}. Both simulators initialize a user goal when the dialog starts\footnote{Refer to the appendix for user goal generation.}, provide the agent with a simulated user response at each dialog turn, and work at the dialog act level. Since the original corpus only annotates the dialog acts at the system side, we use the annotation at the user side from ConvLab \cite{lee2019convlab} to implement the two simulators.

\subsection{Evaluation Metrics}
Evaluation of a task-oriented dialog mainly consists of the cost (dialog turns) and task success (inform F1 \& match rate). The definition of inform F1 and match rate is explained as follows.
\paragraph{Inform F1}: This evaluates whether all the \textbf{requested information} (e.g. address, phone number of a hotel) has been informed. Here we compute the F1 score so that a policy which greedily answers all the attributes of an entity will only get a high \textit{recall} but a low \textit{precision}.

\paragraph{Match rate}: This evaluates whether the booked entities match all the \textbf{indicated constraints} (e.g. Japanese food in the center of the city) for all domains. If the agent fails to book an entity in one domain, it will obtain 0 score on that domain. This metric ranges from 0 to 1 for each domain, and the average on all domains stands for the score of a session.

Finally, a dialog is considered successful only if all the information is provided (i.e. inform recall = 1) and the entities are correctly booked (i.e. match rate = 1) as well\footnote{If the user does not request any information in the session, this will just compute match rate, and similarly for inform recall.}.
Dialog success is either 0 or 1 for each session.

\subsection{Implementation Details}

\begin{table}[!tb]
    \centering
    \small
    \begin{tabular}{cc}
    \toprule
        Hyper-parameter & Value \\
    \midrule
        Learning rate & 1e-4 \\
        Mini-batch size & 32 \\
        Discount factor $\gamma$ & 0.99 \\
        Clipping factor $\epsilon$ in PPO & 0.2 \\
        GAE factor $\lambda$ in PPO & 0.95 \\
    \bottomrule
    \end{tabular}
    \caption{Hyper-parameter settings.}
    \label{tab:hyper}
\end{table}

Both the dialog policy $\pi(a|s)$ and the value function $V(s)$ are implemented with two hidden layer MLPs. For the reward estimator $f(s,a)$, it is split into two networks $g(s,a)$ and $h(s)$ according to the proposed algorithm, where each is a one hidden layer MLP. The activation function is all \textit{Relu} for MLPs. We use Adam as the optimization algorithm. The hyper-parameters of GDPL used in our experiments are shown in Table \ref{tab:hyper}.

\subsection{Baselines}
First of all, we introduce three baselines that use handcrafted reward functions. Following \cite{peng2017composite}, the agent receives a positive reward of $2L$ for success at the end of each dialog, or a negative reward of $-L$ for failure, where $L$ is the maximum number of turns in each dialog and is set to $40$ in our experiments. Furthermore, the agent receives a reward of $-1$ at each turn so that a shorter dialog is encouraged.

\paragraph{GP-MBCM} \cite{gavsic2015policy}: Multi-domain Bayesian Committee Machine for dialog management based on Gaussian process, which decomposes the dialog policy into several domain-specific policies.
\paragraph{ACER} \cite{wang2017sample}: Actor-Critic RL policy with Experience Replay, a sample efficient learning algorithm that has low variance and scales well with large discrete action spaces.
\paragraph{PPO} \cite{schulman2017proximal}: The same as the dialog policy in GDPL.

Then, we also compare with another strong baseline that involves reward learning.
\paragraph{ALDM} \cite{liu2018adversarial}: Adversarial Learning Dialog Model that learns dialog rewards with a Bi-LSTM encoding the dialog sequence as the discriminator to predict the task success. The reward is only estimated at the end of the session and is further used to optimize the dialog policy.

For a fair comparison, each method is pre-trained for 5 epoches by simple imitation learning on the state-action pairs.

\section{Result Analysis}

\begin{figure*}[!tb]
    \centering
    \includegraphics[width=0.32\linewidth]{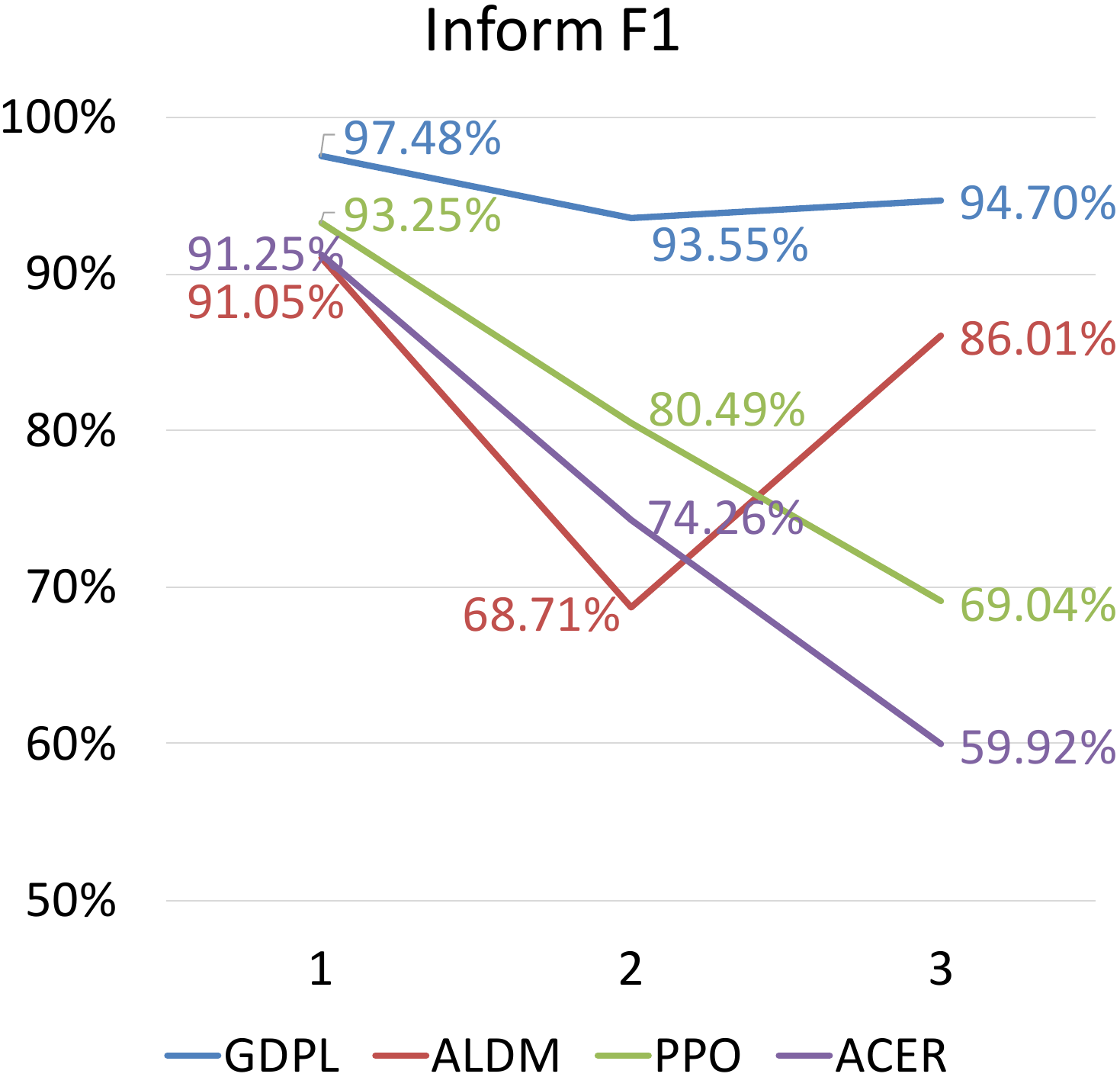}
    \includegraphics[width=0.32\linewidth]{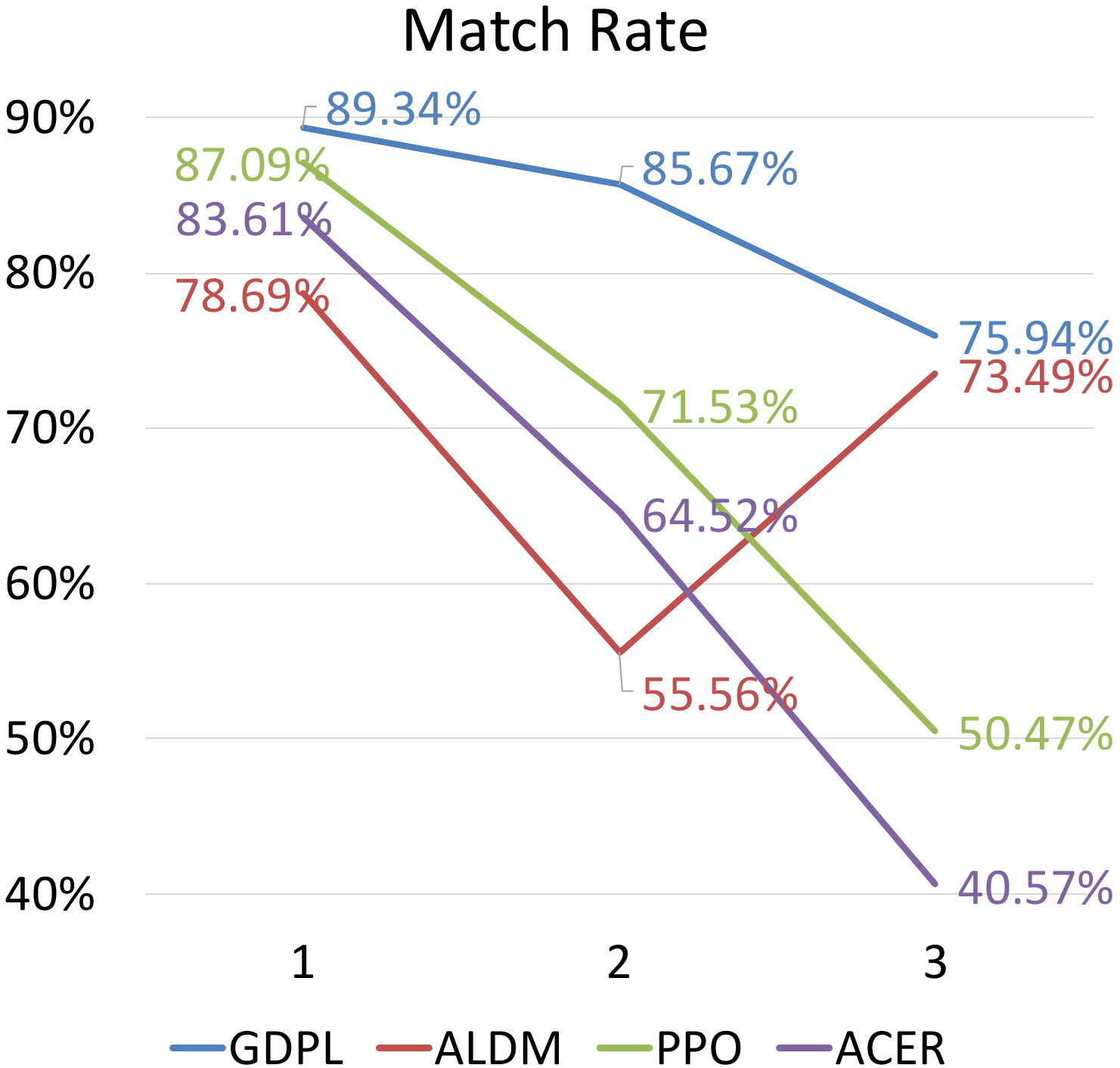}
    \includegraphics[width=0.32\linewidth]{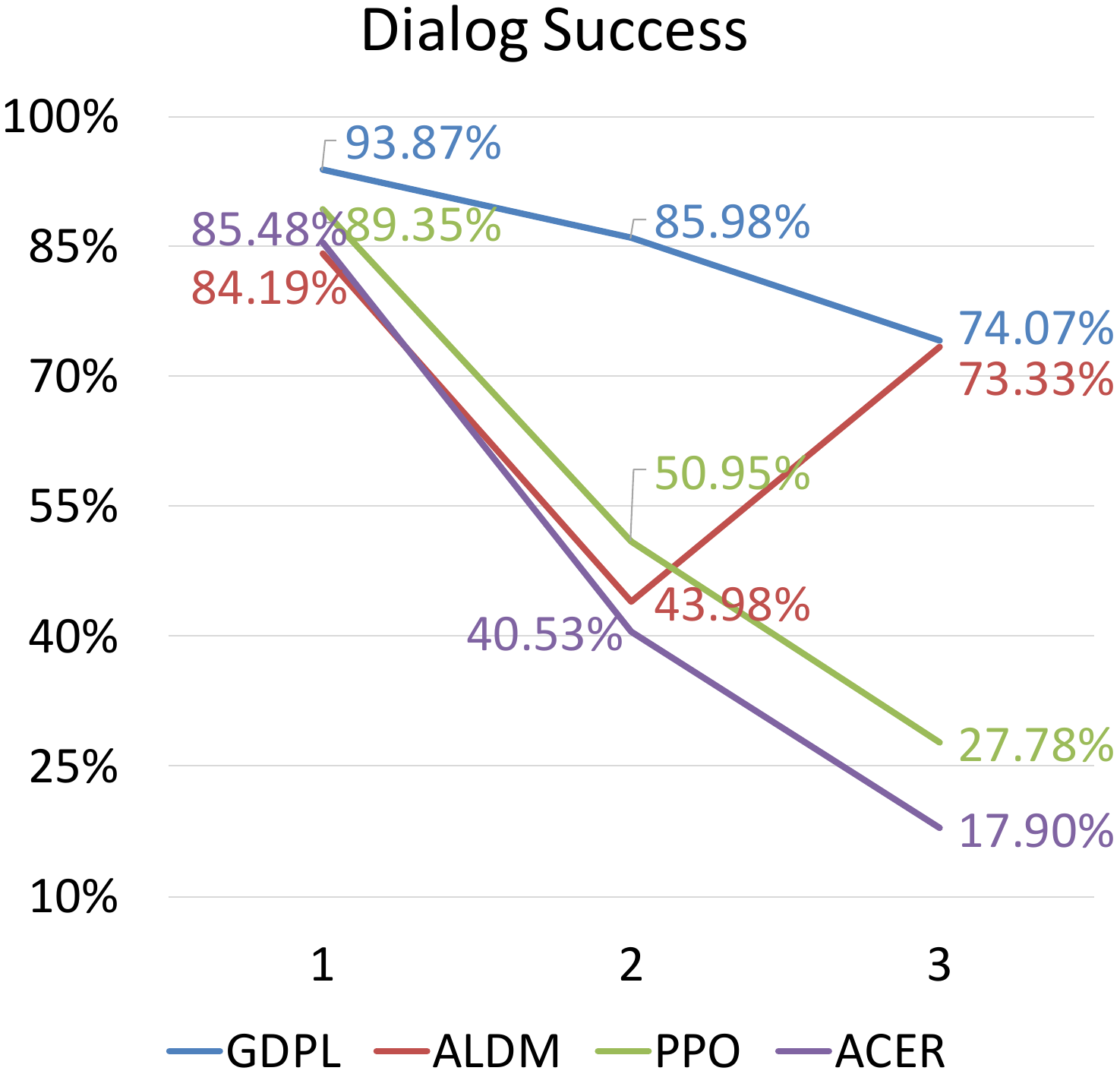}
    \caption{Performance of dialog agents according to the different number of domains in the dialog session. The ratio of the sessions with 1:2:3 domains is 310:528:162 respectively.}
    \label{fig:domain}
\end{figure*}

\subsection{Main Results}

\begin{table}[!tb]
\centering
\small
    \begin{tabular}{lcccccccc}
    \toprule
    \multirow{2}{*}[-0.04in]{Method} & \multicolumn{4}{c}{Agenda} \\
    \cmidrule(lr){2-5}
    & Turns & Inform & Match & Success\\
    \midrule
    GP-MBCM & 2.99 & 19.04 & 44.29 & 28.9 \\
    ACER & 10.49 & 77.98 & 62.83 & 50.8 \\
    PPO & 9.83 & 83.34 & 69.09 & 59.1 \\
    ALDM & 12.47 & 81.20 & 62.60 & 61.2 \\
    \midrule
    GDPL-sess & \textbf{7.49} & 88.39 & 77.56 & 76.4 \\
    GDPL-discr & 7.86 & 93.21 & 80.43 & 80.5 \\
    GDPL & 7.64 & \textbf{94.97} & \textbf{83.90} & \textbf{86.5}\\
    \midrule
    \textit{Human} & \textit{7.37} & \textit{66.89} & \textit{95.29} & \textit{75.0} \\
    \bottomrule
    \end{tabular}
\caption{Performance of different dialog agents on the multi-domain dialog corpus by interacting with the agenda-based user simulator. All the results except ``dialog turns'' are shown in percentage terms. Real human-human performance computed from the test set (i.e. the last row) serves as the upper bounds.}
\label{tab:agenda}
\end{table}

The performance of each approach that interacts with the agenda-based user simulator is shown in Table \ref{tab:agenda}.
GDPL achieves extremely high performance in the task success on account of the substantial improvement in inform F1 and match rate over the baselines. Since the reward estimator of GDPL evaluates state-action pairs, it can always guide the dialog policy during the conversation thus leading the dialog policy to a successful strategy, which also indirectly demonstrates that the reward estimator has learned a reasonable reward at each dialog turn.
Surprisingly, GDPL even outperforms human in completing the task, and its average dialog turns are close to those of humans, though GDPL is inferior in terms of match rate. Humans almost manage to make a reservation in each session, which contributes to high task success. However, it is also interesting to find that human have low inform F1, and that may explain why the task is not always completed successfully. Actually, there have high recall (\textit{86.75\%}) but low precision (\textit{54.43\%}) in human dialogs when answering the requested information. This is possibly because during data collection human users forget to ask for all required information of the task, as reported in \cite{su2016line}.

ACER and PPO obtain high performance in inform F1 and match rate as well. However, they obtain poor performance on the overall task success, even when they are provided with the designed reward that already knows the real user goals. This is because they only receive the reward about the success at the last turn and fail to understand what the user needs or detect the change of user goals.

Though ALDM obtains a lower inform F1 and match rate than PPO, it gets a slight improvement on task success by encoding the entire session in its reward estimator. This demonstrates that learning effective rewards can help the policy to capture user intent shift, but the reward sparsity issue remains unsolved. This may explain why the gain is limited, and ALDM even has longer dialog turns than others. In conclusion, the dialog policy benefits from the guidance of the reward estimator per dialog turn.

\begin{table}[!tb]
    \centering
    \small
    \begin{tabular}{ccccc}
    \toprule
        GP-MBCM & ACER & PPO & ALDM & GDPL \\
    \midrule
        1.666 & 0.775 & 0.639 & 1.069 & \textbf{0.238} \\
    \bottomrule
    \end{tabular}
    \caption{KL-divergence between different dialog policy and the human dialog $KL(\pi_{turns}||p_{turns})$, where $\pi_{turns}$ denotes the discrete distribution over the number of dialog turns of simulated sessions between the policy $\pi$ and the agenda-based user simulator, and $p_{turns}$ for the real human-human dialog.}
    \label{tab:turn}
\end{table}

Moreover, GDPL can establish an efficient dialog thanks to the learned rewards that infer human behaviors. Table \ref{tab:turn} shows that GDPL has the smallest KL-divergence to the human on the number of dialog turns over the baselines, which implies that GDPL behaves more like the human. It seems that all the approaches generate many more \textit{short dialogs} (dialog turns less than 3) than human, but GDPL generates far less \textit{long dialogs} (dialog turns larger than 11) than other baselines except GP-MBCM. Most of the long dialog sessions fail to reach a task success.

We also observe that GP-MBCM tries to provide many dialog acts to avoid the negative penalty at each turn, which results in a very low inform F1 and short dialog turns. However, as explained in the introduction, a shorter dialog is not always the best. The dialog generated by GP-MBCM is too short to complete the task successfully. GP-MBCM is a typical case that focuses too much on the cost of the dialog due to the handcrafted reward function and fails to realize the true target that helps the users to accomplish their goals.

\begin{table}[!tb]
    \centering
    \small
    \begin{tabular}{lcccc}
    \toprule
    \multirow{2}{*}[-0.04in]{Method} & \multicolumn{4}{c}{VHUS} \\
    \cmidrule(lr){2-5}
    & Turns & Inform & Match & Success\\
    \midrule
    ACER & 22.35 & 55.13  & 33.08 & 18.6 \\
    PPO & \textbf{19.23} & \textbf{56.31} & 33.08 & 18.3 \\
    ALDM & 26.90 & 54.37 & 24.15 & 16.4 \\
    \midrule
    GDPL & 22.43 & 52.58 & \textbf{36.21} & \textbf{19.7} \\
    \bottomrule
    \end{tabular}
    \caption{Performance of different agents on the neural user simulator.}
    \label{tab:vhus}
\end{table}

\subsection{Ablation Study}
Ablation test is investigated in Table \ref{tab:agenda}. GDPL-sess sums up all the rewards at each turn to the last turn and does not give any other reward before the dialog terminates, while GDPL-discr is to use the discriminator form as \cite{fu2018learning} in the reward estimator. It is perceptible that GDPL has better performance than GDPL-sess on the task success and is comparable regarding the dialog turns, so it can be concluded that GDPL does benefit from the guidance of the reward estimator at each dialog turn, and well addresses the reward sparsity issue. GDPL also outperforms GDPL-discr which means directly optimizing $f_\omega$ improves the stability of AL.

\subsection{Interaction with Neural Simulator}

The performance that the agent interacts with VHUS is presented in Table \ref{tab:vhus}. VHUS has poor performance on multi-domain dialog. It sometimes becomes insensible about the dialog act so it often gives unreasonable responses. Therefore, it is more laborious for the dialog policy to learn a proper strategy with the neural user simulator. All the methods cause a significant drop in performance when interacting with VHUS. ALDM even gets worse performance than ACER and PPO. In comparison, GDPL is still comparable with ACER and PPO, obtains a better match rate, and even achieves higher task success. This indicates that GDPL has learned a more robust reward function than ALDM.

\subsection{Goal across Multiple Domains}

Fig. \ref{fig:domain} shows the performance with the different number of domains in the user goal. In comparison with other approaches, GDPL is more scalable to the number of domains and achieves the best performance in all metrics. PPO suffers from the increasing number of the domain and has remarkable drops in all metrics. This demonstrates the limited capability for the handcrafted reward function to handle complex tasks across multiple domains in the dialog.

ALDM also has a serious performance degradation with 2 domains, but it is interesting to find that ALDM performs much better with 3 domains than with 2 domains. We further observe that ALDM performs well on the taxi domain, most of which appear in the dialogs with 3 domains. Taxi domain has the least slots for constraints and requests, which makes it easier to learn a reward about that domain, thus leading ALDM to a local optimal. In general, our reward estimator has higher effectiveness and scalability.

\subsection{Human Evaluation}

\begin{table}[!tb]
    \centering
    \small
    \begin{tabular}{lccccccccc}
    \toprule
        \multirow{2}{*}[-0.04in]{VS.} & \multicolumn{3}{c}{Efficiency} & \multicolumn{3}{c}{Quality} & \multicolumn{3}{c}{Success} \\
        \cmidrule(lr){2-4} \cmidrule(lr){5-7} \cmidrule(lr){8-10}
        & W & D & L & W & D & L & W & D & L\\
    \midrule
        ACER & 55&25&20 & 44&32&24 & 52&30&18 \\
        PPO & 74&13&13 & 56&26&18 & 59&31&10 \\
        ALDM & 69&19&12 & 49&25&26 & 61&24&15 \\
    \bottomrule
    \end{tabular}
    \caption{The count of human preference on dialog session pairs that GDPL wins (W), draws with (D) or loses to (L) other methods based on different criteria. One method wins the other if the majority prefer the former one.}
    \label{tab:human}
\end{table}
For human evaluation, we hire Amazon Mechanical Turkers to state their preferences between GDPL and other methods. Because all the policies work at dialog act level, we generate the texts from dialog acts using hand-crafted templates to make the dialog readable. Given a certain user goal, Turkers first read two simulated dialog sessions, one from the interaction between GDPL and the agenda-based user simulator, the other from another baseline with the same simulator. Then, they are asked to judge which dialog is better (win, draw or lose) according to different subjective assessments. In addition to \textit{Task Success}, we examine another two measures concerning \textit{Dialog Cost} in the human evaluation: \textit{Efficiency} such as dialog turn cost or response delay, and \textit{Quality} such as redundant information or inappropriate reply \cite{walker1997paradise}. Since the text is generated by templates for all policies, we do not evaluate language generation here (including grammar, diversity, etc.). We randomly sample 300 user goals from the test set, 100 each for one baseline, and each session pair is evaluated by 3 Turkers.

Table \ref{tab:human} presents the results of human evaluation. GDPL outperforms three baselines significantly in all aspects (sign test, p-value $<$ 0.01) except for the quality compared with ACER. Among all the baselines, GDPL obtains the most preference against PPO. Note that the difference between PPO and GDPL is only in the reward signal. This again demonstrates the advantage of reward learning in GDPL over the handcrafted reward function.
The agreement on the superiority of GDPL between objective rating in Table \ref{tab:agenda} and human preference here also indicates that the automatic metrics used in our experiments is reliable to reflect user satisfaction to some extent.

\begin{figure*}[!tb]
    \centering
    \includegraphics[width=0.8\linewidth]{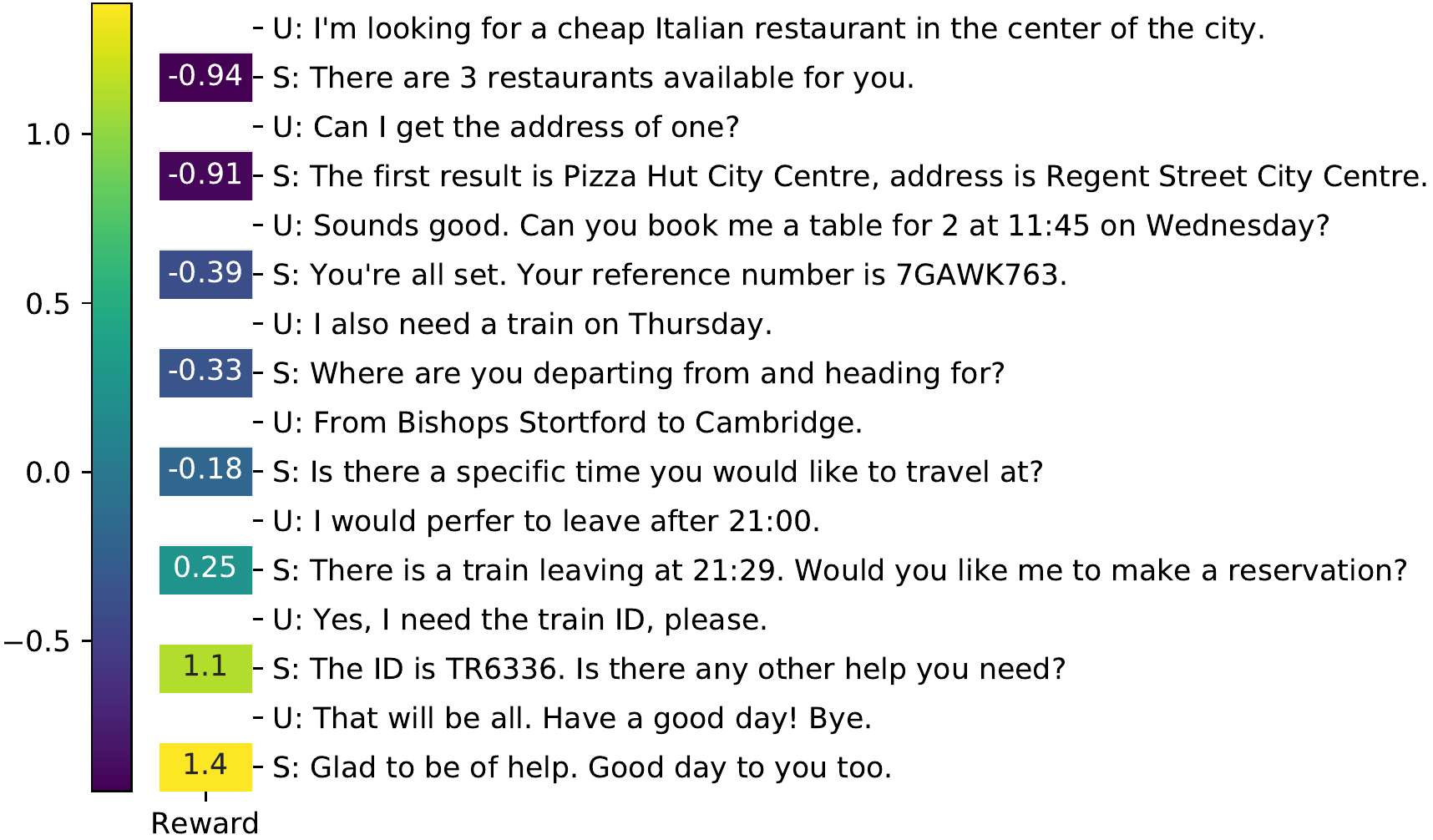}
    \caption{Reward visualization of a dialog session simulated between GDPL and the agenda-based user simulator that contains restaurant and train domains.\footnotemark}
    \label{fig:case}
\end{figure*}

\subsection{Reward Evaluation}

\begin{table}[!tb]
    \centering
    \small
    \begin{tabular}{lcccccc}
    \toprule
    \multirow{2}{*}[-0.04in]{Type} & \multicolumn{2}{c}{Inform} & \multicolumn{2}{c}{Match} & \multicolumn{2}{c}{Success} \\
    \cmidrule(lr){2-3} \cmidrule(lr){4-5} \cmidrule(lr){6-7}
    & Mean & Num & Mean & Num & Mean & Num \\
    \midrule
     Full & 8.413 & 903 & 10.59 & 450 & 11.18 & 865 \\
     Other & -99.95 & 76 & -48.15 & 99 & -71.62 & 135 \\
    \bottomrule
    \end{tabular}
    \caption{Return distribution of GDPL on each metric. The first row counts the dialog sessions that get the full score of the corresponding metric, and the results of the rest sessions are included in the second row.}
    \label{tab:reward}
\end{table}

To provide an insight into the learned reward function itself, Table \ref{tab:reward} provides a quantitative evaluation on the learned rewards by showing the distribution of the return $R = \sum_t \gamma^t r_t$ according to each metric.
It should be noted that some sessions do not have inform F1 because the user does not request any information, and similarly for match rate.
It can be observed that the learned reward function has good interpretability in that the reward is positive when the dialog gets a full score on each metric, and negative otherwise.

Fig. \ref{fig:case} gives an illustration of the learned reward at each dialog turn as a qualitative evaluation. In the beginning, the agent is unaware of the user goal thus it starts with a low reward. As the dialog proceeds, the agent has collected enough information from the user, then books the restaurant successfully and the reward remarkably increases at the third turn. The reward continues to grow stably after the topic shifts to the train domain. Again, the agent offers the correct train ID given sufficient information. Since the user has been informed all the requested information and the restaurant and train are both booked successfully, the user leaves the session with satisfaction at last, and the reward rises to the top as well. In brief, the learned reward can well reflect the current state of the dialog.
It is also noticeable that the dialog policy manages to express multiple intents during the session.

\footnotetext{Refer to the appendix for the dialog acts.}

\section{Discussion}
In this paper, we propose a guided policy learning method for joint reward estimation and policy optimization in multi-domain task-oriented dialog. The method is based on Adversarial Inverse Reinforcement Learning.
Extensive experiments demonstrate the effectiveness of our proposed approach and that it can achieve higher task success and better user satisfaction than state-of-the-art baselines.

Though the action space $\mathcal{A}$ of the dialog policy is defined as the set of all dialog acts, it should be noted that GDPL can be equipped with NLU modules that identify the dialog acts expressed in utterance, and with NLG modules that generate utterances from dialog acts. In this way, we can construct the framework in an end-to-end scenario.

The agenda-based user simulator is powerful to provide a simulated interaction for the dialog policy learning, however, it needs careful design and is lack of generalization. While training a neural user simulator is quite challenging due to the high diversity of user modeling and the difficulty of defining a proper reward function, GDPL may offer some solutions for multi-agent dialog policy learning where the user is regarded as another agent and trained with the system agent simultaneously. We leave this as the future work.

\section*{Acknowledgement}
This work was supported by the National Science Foundation of China (Grant No. 61936010 / 61876096) and the National Key R\&D Program of China (Grant No. 2018YFC0830200). We would like to thank THUNUS NExT Joint-Lab for the support, anonymous reviewers for their valuable suggestions, and our lab mate Qi Zhu for helpful discussions.
The code is available at \url{https://github.com/truthless11/GDPL}.

\bibliography{emnlp-ijcnlp-2019}
\bibliographystyle{acl_natbib}

\newpage
\appendix

\section{User Goal}
In the task-oriented dialog setting, the entire conversation is around a user goal $G=(C,R)$ implicitly, where $C$ denotes the constraint and $R$ is the requests \cite{schatzmann2007agenda}. The user goals appeared in the original corpus are all extracted into one database. Note that each user goal in MultiWOZ \cite{budzianowski2018multiwoz} may consist of a ``real'' user goal that describes what the user wants in the end, along with a different ``failed'' user goal that indicates what the user wants at first. Every time a dialog is launched, the user goal is initialized by the user simulator at the beginning of a dialog session, by randomly sampling the \textit{constraint slots} and \textit{requests slots} from the user goal database. Each slot is sampled according to its frequency in the dataset, and a ``failed'' user goal would be also initialized with a certain probability to simulate the changes in a user goal.

\section{Case Study}
Table \ref{tab:sample} shows two dialog sessions with the same user goal in PPO and GDPL, respectively. The user goal consists of \textit{attraction}, \textit{hotel} and \textit{taxi} domains. The system is required to answer all the information in \textit{Reqt} based on the constraints in \textit{Info} and book a right entity that meets the needs in \textit{Book}. When the user asks for some information about the museum, GDPL manages to provide all the requested information, while PPO only answers part of the information, which causes the user to inquire about entrance fee again (red line). Additionally, GDPL answers the area of the hotel (blue line) and the dialog proceeds successfully, however, PPO fails.

\begin{table*}[!htp] 
\small
\begin{tabular}{c|c|c|c|c|c|c}
\hline
\multicolumn{2}{c|}{\textbf{Attraction}} & \multicolumn{3}{c|}{\textbf{Hotel}} & \multicolumn{2}{c}{\textbf{Taxi}} \\
\hline
Info & Reqt & Info & Reqt & Book & Info & Reqt \\
\hline
type=museum & entrance fee & price range=expensive & address & people=8 & arrive by=19:45 & car type\\
area=centre & address & stars=4 & area & day=saturday & & phone \\
& postcode & type=hotel & & stay=5 & & \\
& & parking=yes & & & & \\
\hline
\end{tabular}
\begin{tabular}{l@{~}l|l@{~}l}
\hline
\multicolumn{2}{c|}{\textbf{PPO}}  & \multicolumn{2}{c}{\textbf{GDPL}} \\
\hline
U:& Hi, I would like to visit a museum in the centre area. & U:& Hi, I would like to visit a museum in the centre area.  \\
S:& I found 1 attraction matching your requirements. & S:& The \textit{regency gallery} matches your requirements.  \\
U:& May I have its address, postcode, and entrance fee? & U:& May I have its address, postcode, and entrance fee?  \\
S:& The address of the \textit{regency gallery} is \textit{the grafton}  & S:& Its address is \textit{the grafton centre, east road}, \\
  & \textit{centre, east road}, and postcode is \textit{cb11ps}. & & postcode is \textit{cb11ps}, and admission is free.\\
U:& \textcolor{red}{\underline{What about the entrance fee?}} & U:& I also need to book an expensive 4-star hotel \\
  & & & that has parking.\\
S:& Admission is free. & S:& There are 2 hotels matching your requirements.  \\
U:& I also need to book an expensive 4-star hotel & U:& Could you tell me the address and area of one? \\
  & that has parking. & & \\
S:& There are 2 hotels matching your requirements. &  S:& The address of \textit{huntingdon marriott hotel} is \textit{kingfisher} \\
  & Its name is \textit{huntingdon marriott hotel}. & & \textit{way, hinchinbrook business park, huntingdon}. \\
U:& Could you tell me the address and area of one? & U:& \textcolor{blue}{\underline{What is the area of the hotel?}}  \\
S:& The address of \textit{huntingdon marriott hotel} is \textit{kingfisher} & S:& It's located on the centre of city. \\
  & \textit{way, hinchinbrook business park, huntingdon}. & & \\
U:& \textcolor{blue}{\underline{What is the area of the hotel?}} &  U:& Please book it for 8 people and 5 nights on saturday. \\
S:& The name is \textit{huntingdon marriott hotel}. &  S:& Your reference number is 5NAWGJDC.   \\
U:& \textcolor{blue}{\underline{What is the area of the hotel?}} & U:& Could you also book me a taxi that arrives by 19:45? \\
S:& The name is \textit{huntingdon marriott hotel}. & S:& Booking is successful. It will be a yellow honda \\
  & & & and the phone number is 3736171789. \\
U:& \textcolor{blue}{\underline{What is the area of the hotel?}} & U:& Thank you. That will be all. \\
S:& The name is \textit{huntingdon marriott hotel}. & S:& You are welcome. Goodbye. \\
\multicolumn{2}{c|}{\textbf{\vdots}} &  \\
U:& \textcolor{blue}{\underline{What is the area of the hotel?}}&   \\
S:& The name is \textit{huntingdon marriott hotel}. &   \\
\hline
\multicolumn{2}{c|}{\textbf{Failure:} \begin{tabular}[c]{@{}l@{}}Agent does not tell the area of the hotel properly.\end{tabular}} & \multicolumn{2}{c}{\textbf{Success}}
\\ \hline
\end{tabular}
\small
\centering
\caption{A sample dialog session comparison between PPO and GDPL.}
\label{tab:sample}
\end{table*}

\section{Reward Visualization}
\begin{figure*}[!tbp]
    \centering
    \includegraphics[width=\linewidth]{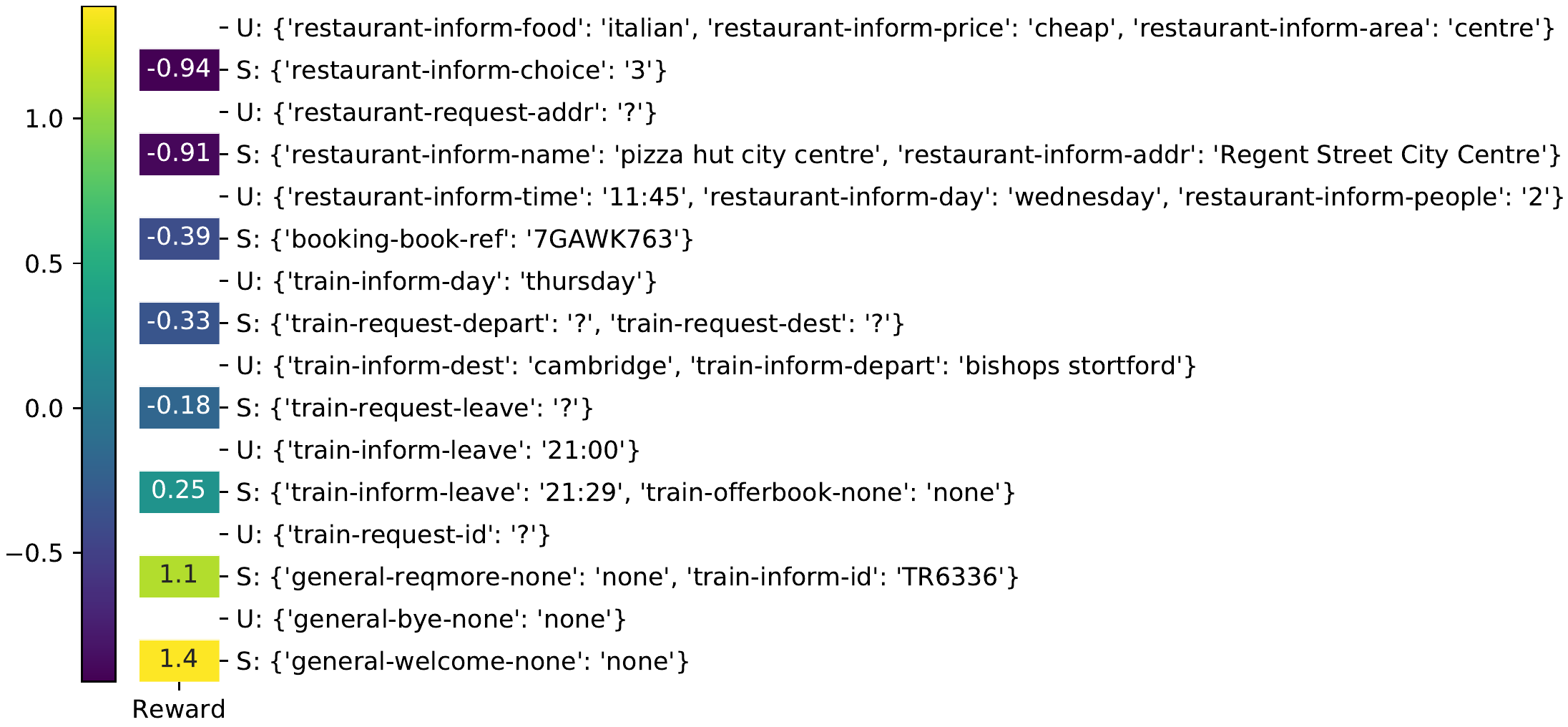}
    \caption{Reward visualization with dialog acts.}
    \label{fig:reward_da}
\end{figure*}

Fig. \ref{fig:reward_da} shows the reward visualization of the session discussed in the \textit{reward evaluation} subsection with its original dialog acts.

\end{document}